\documentclass[sigconf]{acmart}
\usepackage{xcolor,colortbl}
\usepackage{booktabs} % For formal tables
\usepackage{todonotes}
\usepackage{caption}
\usepackage{subcaption}

\setcopyright{rightsretained}
\acmConference[HT'2018]{ACM HT}{July 2018}{Baltimore, Maryland} 
\acmYear{2018}
\copyrightyear{2018}

\begin{document}
%\title{Show Me Your Friends and I'll Tell You Who You Are: Predicting Occupation and Income Using User's Social Network}
\title{Predicting Twitter User Socioeconomic Attributes with Network and Language Information}
%\titlenote{Produces the permission block, and copyright information}
% \subtitle{Extended Abstract}
% \subtitlenote{The full version of the author's guide is available as
%   \texttt{acmart.pdf} document}

\author{Nikolaos Aletras}
% \authornote{Dr.~Trovato insisted his name be first.}
\affiliation{%
  \institution{University of Sheffield}
}
\email{n.aletras@sheffield.ac.uk}

\author{Benjamin Paul Chamberlain}
% \authornote{The secretary disavows any knowledge of this author's actions.}
\affiliation{%
  \institution{Imperial College London}
}
\affiliation{%
  \institution{Asos.com}
}
\email{benjamin.chamberlain@gmail.com}

\renewcommand{\shortauthors}{Aletras and Chamberlain}

\begin{abstract}
Inferring socioeconomic attributes of social media users such as occupation and income is an important problem in computational social science. Automated inference of such characteristics has applications in personalised recommender systems, targeted computational advertising and online political campaigning. While previous work has shown that language features can reliably predict socioeconomic attributes on Twitter, employing information coming from users' social networks has not yet been explored for such complex user characteristics. In this paper, we describe a method for predicting the occupational class and the income of Twitter users given information extracted from their extended networks by learning a low-dimensional vector representation of users, i.e. graph embeddings. We use this representation to train predictive models for occupational class and income. Results on two publicly available datasets show that our method consistently outperforms the state-of-the-art methods in both tasks. We also obtain further significant improvements when we combine graph embeddings with textual features, demonstrating that social network and language information are complementary.
\end{abstract}



% We no longer use \terms command

\keywords{social media, graph embeddings, user profiling}

%% Used in some conference proceedings e.g. sigplan and sigchi
%   \caption{This is a teaser}

\maketitle

\section{Introduction}
\label{sec:intro}

% why inferring user attributes is important?

The daily interaction of billions of users with online social platforms such as Facebook and Twitter has made available enormous amounts of user generated content. The plethora and diversity of this data (e.g. text, images or interactions with other users such as `retweets' or `likes') enables studies in computational social science \cite{Lazer2009,Conte2012} to analyse human behaviour on a large scale and automatically infer user latent attributes. 

Automatic inference of user characteristics %involves multidisciplinary efforts in the areas of data science, natural language processing, information retrieval and social sciences. It and 
includes studies on inferring age and gender \cite{Rao2010,Burger2011,Schwartz2013}, location \cite{Cheng2010,Han2014,Dredze2016}, personality traits \cite{Quercia2011,Kosinski2013,Volkova2015} and political orientation \cite{Tumasjan2010,Pennacchiotti2011,Cohen2013} inter alia. More recently, there has been a particular focus on inferring complex user socioeconomic characteristics such as occupational class \cite{Li2014,Huang2015,jobs2015}, income \cite{income2015,Volkova2015} and socioeconomic class \cite{Filho2014,Lampos2016}. Apart from their importance in computational social science, such methods are also useful in downstream applications such as targeted advertising and online political campaigning.%, where political parties and candidates can target audiences with specific characteristics.

% what are the limitations of previous work?
% links to social science
Following the hypothesis that language is indicative of the social status of a person \cite{Bernstein1960,Bernstein2003,Labov2006}, previous research analysed user generated written content to derive text based features such as bag-of-words or clusters of words. These features are used to train predictive models for inferring socioeconomic attributes \cite{Li2014,jobs2015}. Despite the fact that these methods have proved to perform well, they have not considered any relations and interactions between users. Moreover, there is a large proportion of inactive users that do not produce any content. For example, previous studies have shown that only around two thirds of the users are active (i.e. posted at least twice) on Twitter \cite{Huberman2008,Liu2014}. This makes it impossible to solely utilise language based models to infer socioeconomic or other characteristics of inactive users. 

A different approach to the problem is to include information from the social network structure. Socioeconomic status can be indicated by looking into the range and the composition of the social network of a person \cite{Campbel1986}. That is because people who belong to the same social circles often share common characteristics. This is known as social network homophily, i.e. the inclination of people towards developing social ties with similar others \cite{Lazarsfeld1954,Mcpherson2001}. %People who belong to the same class, may have similar education, ethnic background, or even share common political views and interests. 
Despite expected differences to real life social networks, it has been shown that online social networks, e.g. Facebook and Twitter, exhibit some levels of homophily \cite{Kosinski2013, AlZamal2012}. People that follow each other on Twitter usually share common topical interests \cite{Kwak2011,Weng2010}. Previous work utilised the social network structure to infer user attributes such as gender and age, personality traits and sentiment \cite{Tan2011,AlZamal2012,Staiano2012,Kosinski2013,Perozzi2015}, but not any socioeconomic attributes.

% what do we do?
% what are the main contributions?
In this paper, we focus on using social network information to infer user's occupational class and income. Following that direction, we explore two hypotheses using data from Twitter: (1) a user's social network is indicative of their income and occupational class; and (2) the information from the social network structure and textual information are complementary. To answer these hypotheses, we extract information from a user's social network and encapsulate it in user graph embeddings \cite{Perozzi2014,Tang2015}. Graph embeddings place Twitter users in a vector space where similar users are likely to be close to each other. The user graph embeddings are treated as features to train linear and non-linear supervised models for predicting income and occupational class. 

%Our evaluation on two standard, publicly available datasets of Twitter users that are labelled with occupational class and income shows that the social network information significantly outperforms language based features introduced in previous studies. Furthermore, we find that the combination of information extracted from both sources of data, i.e. social network and language, improves predictive performance beyond what is achieved by a single source.

The major contributions of our paper are:
\begin{itemize}
\item To the best of our knowledge, this work is the first to introduce neural graph embeddings to predict income and occupational class on Twitter. 
\item Our model can be used to infer complex socioeconomic characteristics of inactive users, exploiting the fact that user graph embeddings do not rely on any textual information.
\item We show that a user's social network and written content, i.e. tweets, contain complementary information. This is demonstrated by training models that combine both feature sets. Our evaluation on two standard, publicly available datasets of Twitter users that are labelled with occupational class and income shows that they outperform models using solely language or solely network information.
\item Our proposed model achieves state-of-the-art performance in these two datasets of income and occupational class significantly outperforming models introduced by \citet{jobs2015,income2015}.
\end{itemize}

\section{User Neural Graph Embeddings}
\label{sec:embeddings}
% introduction
User neural graph embeddings are dense vector representations that position similar users close together in a high-dimensional Euclidean space. Neural embeddings are popular in natural language processing for learning vector representations of words \cite{Mikolov2013a,Bengio2003}. %Word embeddings improve performance in a broad range of linguistic tasks such as machine translation \cite{Zou2013} and sentiment analysis \cite{Tang2014}. 

% extension to graphs
The only inputs required to learn word embedding models are sequences of words in documents and so the concept can be extended to network structured data using random walks to create sequences of vertices. In our case, vertices represent Twitter users and edges represent a follower/followee relationship, and we treat edges as if they were undirected. This is justified because Twitter is predominantly an interest graph~\citep{Gupta2013} and a large body of research has shown that the homophily principle applies to users who express similar interests in social networks \cite{Chamberlain2017,Kosinski2013,Tan2011}. By treating the graph as undirected we ensure that all users that follow a common account (indicative of an interest) have a maximum path distance of two. %We train embeddings based on the vertex sequences produced from random walks over the undirected graph. 
Vertices are embedded by treating them exactly analogously to words in the text formulation of the model \cite{Perozzi2014}. Extensions varying the nature of the random walks have been explored in LINE \cite{Tang2015} and Node2vec \cite{Grover2016}. The main justification for this idea is that social networks are a form of noisy measurement of a true underlying network. Random walks have been shown extensively to mitigate for false edges and infer the presence of missing ones \cite{Page1999}.

\subsection{Generating User Sequences and Contexts}

% \begin{figure}
% \centering
% \includegraphics[width=0.2\textwidth]{zachary_karate}
% \caption{The network diagram for Zachary's famous karate club. %It is split into two factions centred around vertices 1 (the karate instructor) and 34 (the club president)
% }
% \label{fig:karate_club}
% \end{figure}

\begin{figure}
\centering
\includegraphics[width=0.45\textwidth]{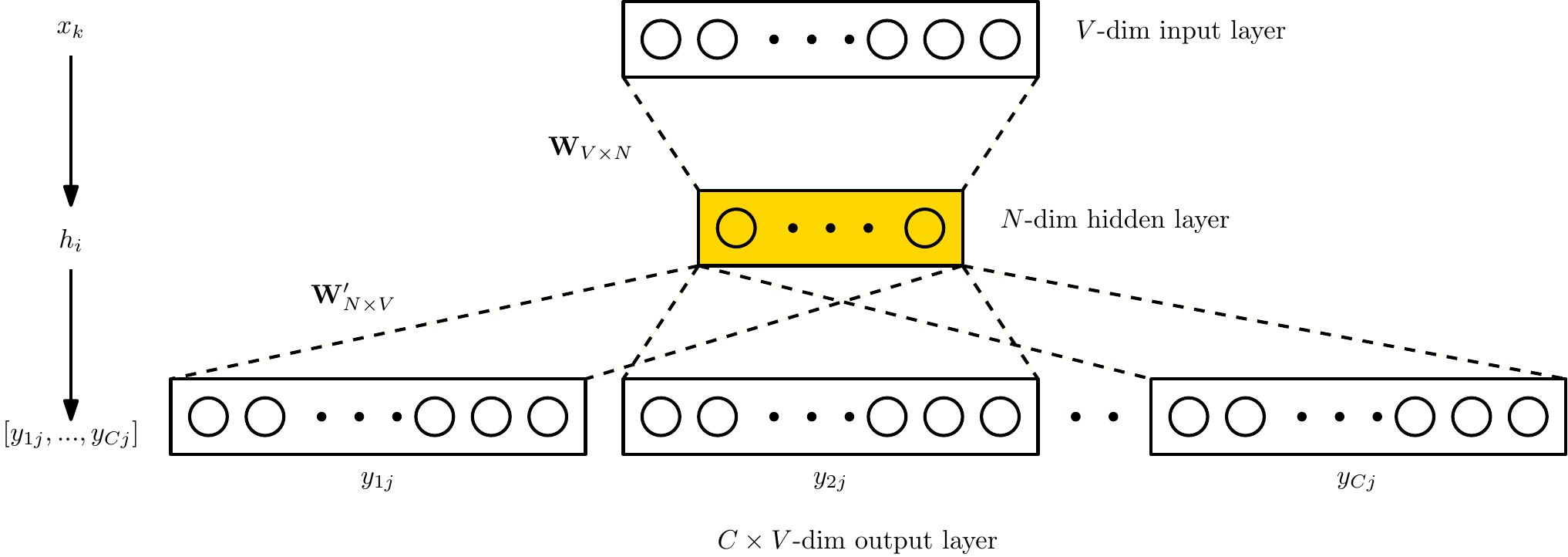}
\caption{The Skipgram model uses two vector representations $\mathbf{W}$ and $\mathbf{W'}$ to predict the context vertices from a single input vertex.}
\label{fig:SGNS}
\end{figure}

Given a network of users connected with unweighted edges, random walks are generated by repeatedly sampling an integer uniformly from \{1,2,\dots, $D_v$\} where $D_v$ is the vertex degree and moving to a new vertex. Concretely, for a random walk starting at vertex $v_0$ we would sample $x \sim U(\{1,2,\dots, D_{v_0}\})$ where U is a uniform distribution and $D_{v_0}$ is the degree of $v_0$. If $x=1$ we move to the lowest indexed neighbour of $v_0$, append that vertex to the random walk and repeat the process at vertex $v_1$.

\subsection{The Skipgram Model on User Sequences}
After we have sampled user sequences with random walks, we can use them to train user embeddings. There are several related embedding models, i.e. SkipGram and Continuous Bag of Words \cite{Mikolov2013a}. Here we adopt the SkipGram with Negative Sampling (SGNS) model that is depicted in Figure~\ref{fig:SGNS}. 
The figure shows a shallow neural network with a single hidden layer and two separate vector representations labelled as \textbf{W} and \textbf{W'}.
The input to SGNS is a sequence of users, which are mapped to (input, context) pairs by sliding a context window over the input sequences. %As an example the word sequence "fortune favors the prepared mind" with a window of size 3 would generate the following (input, context) pairs: (favors, [chance, the]), (the, [favors, prepared]), etc. 
% Returning to the example of the karate network, a typical random walk of length four originating at vertex 27 would be (27, 30, 24, 28). Applying a context window of size three to this walk sequence would generate the following (input, context) pairs: (30,[27,24]) and (24,[30,28]).
The input user representation is in \textbf{W} and the neighbouring (i.e. context) users share a representation in \textbf{W'}. Users are initially randomly allocated within the two vector spaces and then the model is trained using Stochastic Gradient Descent (SGD). The objective function gives the probability of the context users given the input user, which is modelled by a softmax. We optimise the negative log likelihood given by 

{\small
\begin{align}
L &= -\log p(w_{o,1}, w_{o,2}, w_{o,3},...,w_{o,C}|w_I) \\
&= -\log\prod_{c=1}^C\frac{\exp{\mathbf{v}^{'T}_c}\mathbf{v}_I}{\sum_{j=1}^V\exp{\mathbf{v}^{'T}_j \mathbf{v}_I}}
\label{eq:objective}
\end{align}
}
where $w_{(.)}$ is a user and $\mathbf{v}_{(.)}$ and $\mathbf{v}_{(.)}^{'}$ are the input and output vector representations of that user and $C$ is the context size, typically ten.
% negative sampling
In practice, it is expensive to evaluate Equation~\eqref{eq:objective} as the sum in the denominator is over all of the users in the network. Instead we use negative sampling, which is a form of Noise Contrastive Estimation (NCE) \cite{Gutmann2012}, to estimate the function by only evaluating a small number of negative samples in addition to the observed positive example. The gradient descent update rules for a user pair $(w_I, w_{O})$ with vector representations $(\mathbf{v}_I, \mathbf{v}'_O)$ are found by applying the chain rule to Equation~\eqref{eq:objective} and are given by

{\small
\begin{align}
\mathbf{v}_j^{'new} &= 
\begin{cases} 
\mathbf{v}_j^{'old} - \eta(\sigma(\mathbf{v}_j^{'T}\mathbf{v}_I) - t_j)\mathbf{v}_i, &  w_j \in \chi\\ 
\mathbf{v}_j^{'old}, & \text{otherwise} 
\end{cases}
\end{align}
}%
where $\chi = \{w_O\}\cup W_{neg}$. For the output representation and 
\begin{align}
\mathbf{v}_I^{new} &= 
\mathbf{v}_I^{old} - \eta \sum_{j : w_j \in \chi} (\sigma(\mathbf{v}^{'T}_j \mathbf{v}_I) - t_j)\mathbf{v}'_j
\end{align}
for the input representation. In these equations $t_j$ is an indicator variable that is one if and only if $w_j = w_O$ and zero otherwise, $\eta$ is the SGD learning rate and $W_{\rm neg}$ is the set of negatively sampled users. We follow \cite{Mikolov2013a} and draw $W_{\rm neg}$ from the distribution of users in the random walks raised to the power of $\frac{3}{4}$. %The geometric interpretation of the update rules is illustrated in Figure~\ref{fig:sgns_updates} for the output representations v'. 
\section{Experimental Setup}
\label{sec:experimental_setup}

\renewcommand\arraystretch{1.1}
\begin{table}[!t]
\footnotesize
% \tiny
\begin{tabular}{c|l|c}
\hline
\hline
{\bf C}	& {\bf Title} & {\bf U}\\
\hline
1	& Managers, Directors and Senior Officials	&	461 \\
2	&	Professional Occupations &	1615 \\
3	&	Associate Profess. and Technical Occupations &	950 \\
4	&	Administrative and Secretarial Occupations &	168 \\
5	&	Skilled Trades Occupations &	782 \\
6	&	Caring, Leisure and Other Service Occup. &	270 \\
7	&	Sales and Customer Service Occupations &	56 \\
8	&	Process, Plant and Machine Operatives &	192 \\
9	&	Elementary Occupations &	131 \\
\hline
\multicolumn{2}{c|}{Total} & 4625\\
\hline
\hline
\end{tabular}
\caption{Distribution of users (U) across occupational classes (C).}\label{t:jobs}
\end{table}
~
\begin{figure}[!t]
\centering
\includegraphics[width=0.8\linewidth, height=3.5cm]{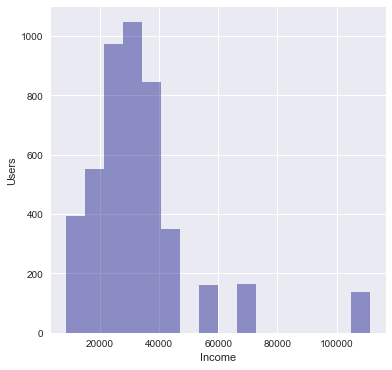}
\caption{Distribution of users and income. Income is calculated in British pounds (\pounds).}
\label{fig:income}
\end{figure}
%This section describes the experimental setup. Specifically, we begin by introducing the datasets used, details on the graph embeddings implementation and the hyperparameter selection of our proposed models. The section finishes with a description of the baselines and the evaluation metrics.

\subsection{Data}
\label{ssec:data}

We experiment using two publicly available datasets that contain Twitter users mapped to their occupational class and income \cite{income2015,jobs2015}. The datasets contain the same group of 5,191 users in total. However, some of the accounts are not considered in our experiments since we were not able to extract their social network information. These accounts may have been deleted or become private since the release of the datasets. Therefore, we report results on a subset of the original set of users, i.e. 4,625 users, that are still publicly available.%\footnote{For more details on how these datasets have been created, refer to \cite{jobs2015,income2015}.}.

\paragraph{Occupational Class} Users are mapped with an occupation using the Standard Occupation Classification (SOC) taxonomy devised by the Office of National Statistics in the UK, based on skill requirements. The SOC taxonomy has a hierarchical structure with 9 major groups (e.g managers or elementary occupations). Users in the dataset have been mapped to one of these major groups. Table~\ref{t:jobs} shows the distribution of users across the nine occupational classes. The Pearson's correlation between the original distribution of users and our subset distribution is $0.93$.

\paragraph{Income}  The occupational class of users has further been used as a proxy to infer their income from the Annual Survey of Hours and Earnings. The income represents the mean yearly earnings for 2013 in British Pounds (GBP) for each occupational class. Figure~\ref{fig:income} shows the distributions of users and income in the dataset. The mean user income in the original dataset is $32,509.74$, while the mean of the subset we use in our experiments is $32,727.92$.

\subsection{Implementation of the Graph Embeddings}

%\paragraph*{User Graph} 
To construct graph embeddings we downloaded the Twitter IDs of everyone followed by the $4,625$ accounts. This produced a set of $3,925,702$ users in total. We considered only accounts followed by at least 10 users, which reduced the number of the unique accounts to $53,199$. To produce sequences of users we treat the edges of the Twitter graph as undirected and take 80 step random walks initiated at each vertex in this network.

%\subsection{On Context Size and Data Epochs}

% \todo[inline]{This paragraph doesn't fit into the paper - its' just an observation I made. It could be replaced with a comment on the effect of the context size now that we have data on this.}
% \paragraph*{Context size and Epochs} The context size determines the order of the random walk that effects the learned embedding. A context size of one only considers directly connected vertices. In previous work, the effect of this parameter has been investigated by fixing the number of data epochs \cite{Grover2016}. This approach however masks a confounding effect. In SGNS each input-output pair is treated independently and as an infinite number of these pairs can be generated by extending the random walks, it is the number of pairs that should be held fixed. As the number of pairs is approximately proportional to the context window, previous approaches do not do this and mask a rapid drop off in embedding quality as the context window grows for a fixed amount of input data.

%\paragraph*{Embedding Size} 
The dimensionality of the embedding affects the performance in predictive tasks. We experimented with dimensionalities of $16, 32, 64$ and $128$ and chose the optimal value following a nested 10-fold cross-validation approach as in \citet{jobs2015,income2015}. We found that the best performing embedding\footnote{During initial experimentation, we noticed that varying the length of the random walk between 40 and 100 did not substantially affect the quality of the embeddings.} dimensionality is 32. The user embeddings and the code to generate them are available to download from \href{https://github.com/melifluos/income-prediction}{{\tt https://github.com/melifluos/income-prediction}}.

\subsection{Predictive Models}
\label{sec:models}

%Following previous work on inferring user socioeconomic characteristics (see Section~\ref{sec:related}), we frame the inference of the occupation as a classification task and the income prediction as a regression task. %We compare state-of-the-art linear and non-linear supervised models introduced in Preo\c{t}iuc-Pietro et al. \cite{jobs2015,income2015} that have been shown to perform well in many user profiling tasks. 

\paragraph*{Occupational Class}
Predicting the occupational class of a user is defined as a 9-way classification task. Given a user feature representation, our goal is to assign the most probable class label. For that purpose, we use the graph embeddings as features and a concatenation of the graph embeddings with the topics introduced in \cite{jobs2015} to train Logistic Regression (LR) \cite{Zou2005}, Support Vector Machines (SVM) \cite{Joachims1998} and Gaussian Process Classifiers (GPC) \cite{Williams1996}. All of the classifiers\footnote{The Gaussian Process models are trained using GPy (\url{http://github.com/SheffieldML/GPy}). All the other models are trained using Scikit-learn (\url{http://scikit-learn.org/}).} 
are trained following the one-vs-all approach\footnote{All the hyperparameters of the baseline predictive models using \emph{Topics} as features are identical to the models presented in \cite{jobs2015,income2015}. We tune the hyperparameters of our proposed models (\emph{Graph} and \emph{Topics+Graph}) performing a nested 10-fold cross-validation, identical to the data splits used in previous work.}.

% \paragraph*{Logistic Regression (LR)} The LR model is a standard logistic regression model with an Elastic Net regularisation. Elastic Net uses a linear combination of the L1 and L2 penalties to regularise the logistic regression model to reduce model complexity and thus overfitting.

% \paragraph*{Support Vector Machine (SVM)} Our SVM model uses a Radial Basis Function (RBF) kernel since that allows us to capture non-linearities within our data.

% \paragraph*{Gaussian Process (GPC)} The GPC is a Bayesian non-parametric Gaussian Process for classification. GPs are generalizations of multivariate Gaussian random variables to infinite sets of random variables. They can also be viewed as a method for defining a prior distribution over functions. For the occupational classification task we use a Sparse GPC \cite{Snelson2006} with an RBF kernel function. The Sparse GPC is parametrised by the lengthscale and the variance as well as the locations of M pseudo-inputs (inducing points), which we learn during training. In our experiments we use 500 inducing points. Our RBF kernel uses Automatic Relevance Determination (ARD) \cite{Neal1996} to learn a separate lengthscale for each feature. The lengthscale determines how close two training examples have to be to influence each other significantly. 

\paragraph*{Income}
Inferring income is defined as a regression task. Given the user feature representation as input, we try to predict a real value representing the user's income. The goal is to minimise the absolute error between the actual and inferred income. We also compare three popular models: (1) linear regression (LR), (2) Support Vector Regression (SVR) \cite{Drucker1997}, and (3) Gaussian Process Regression (GPR) \cite{Williams1996}.

%%%%%%%%%%%%%%%%%%%%%%%%%%%%%%%%%%%%%%%%%%%%%%%%%%%%%%%
\section{Results and Discussion}
\label{sec:results}
%In this section, we present results of our proposed methods and we provide answers to the hypotheses introduced in Section~\ref{sec:intro}.

%\subsection{Performance Comparison}

Tables~\ref{t:results_jobs} and~\ref{t:results_income} show the results obtained by the proposed models using the  graph embeddings (\emph{Graph}) and their combination (\emph{Graph+Topics}) as feature representations for users. Note that models using \emph{Topics} (i.e. word frequency of user's tweets in a set of 200 precomputed word clusters) and \emph{Temporal Orientation} as features are the baseline methods presented in \citet{jobs2015,income2015} and \citet{Hasanuzzaman2017}.\footnote{Replicating the method of \citet{Hasanuzzaman2017} was not possible, hence we report results only for income from their paper.} To compare against \citet{income2015,jobs2015}, we retrain these models using the user accounts in the dataset that are publicly available (see Subsection~\ref{ssec:data}). 

%using topic features, . %The word clusters have been generated using spectral clustering on a similarity matrix of word embeddings trained on a held-out Twitter corpus. 
%We use the same 200 clusters and retrain their models using the user accounts in their dataset that are still publicly available (see Subsection~\ref{ssec:data} above).

\renewcommand\arraystretch{1.0}
\begin{table}[!t]
\centering
\small
\begin{tabular}{|l|c|}
%\hline
\hline
\multicolumn{2}{|c|}{\cellcolor{gray!35}{\bf Occupational Class}}\\
\hline
\cellcolor{gray!35} {\bf Method}	& \cellcolor{gray!35} {\bf Accuracy (\%)} \\
\hline
Majority Class					& 35.00	\\
\hline
\multicolumn{2}{|c|}{\cellcolor{gray!35}{\bf \citet{jobs2015}}}\\ 
\hline
LR-Topics 		& 46.57	\\
SVM-Topics 		& 49.47\\
GP-Topics		& 49.64\\
\hline
\multicolumn{2}{|c|}{\cellcolor{gray!35}{\bf Ours}}\\
\hline
LR-Graph	& 46.24	\\
SVM-Graph	& 50.14	\\
GP-Graph	& 50.44\\
\hline
LR-Graph+Topics		&	48.84\\
SVM-Graph+Topics	&	{\bf 52.00\dag}\\
GP-Graph+Topics	&	51.46\dag \\		
\hline
\end{tabular}
\caption{Accuracy of models in predicting user occupational class.% \emph{Topics} represent topical features, \emph{Graph} represents user graph embeddings, and \emph{Graph+Topics} their combination. 
$\dagger$ denotes statistical significant different (t-test, $p<0.01$) method to \emph{GP-Topics}.} \label{t:results_jobs}
\end{table}

% Both tasks
%%%%%%%%%%%%
% Non-linear models better than the linear ones
% Graph much better than topics in income prediction
% Graph slightly better than topics in job classification
% Graph+Topics further improvements (bigger in job prediction, smaller in income)

\begin{figure}[!t]
\includegraphics[width=0.35\textwidth]{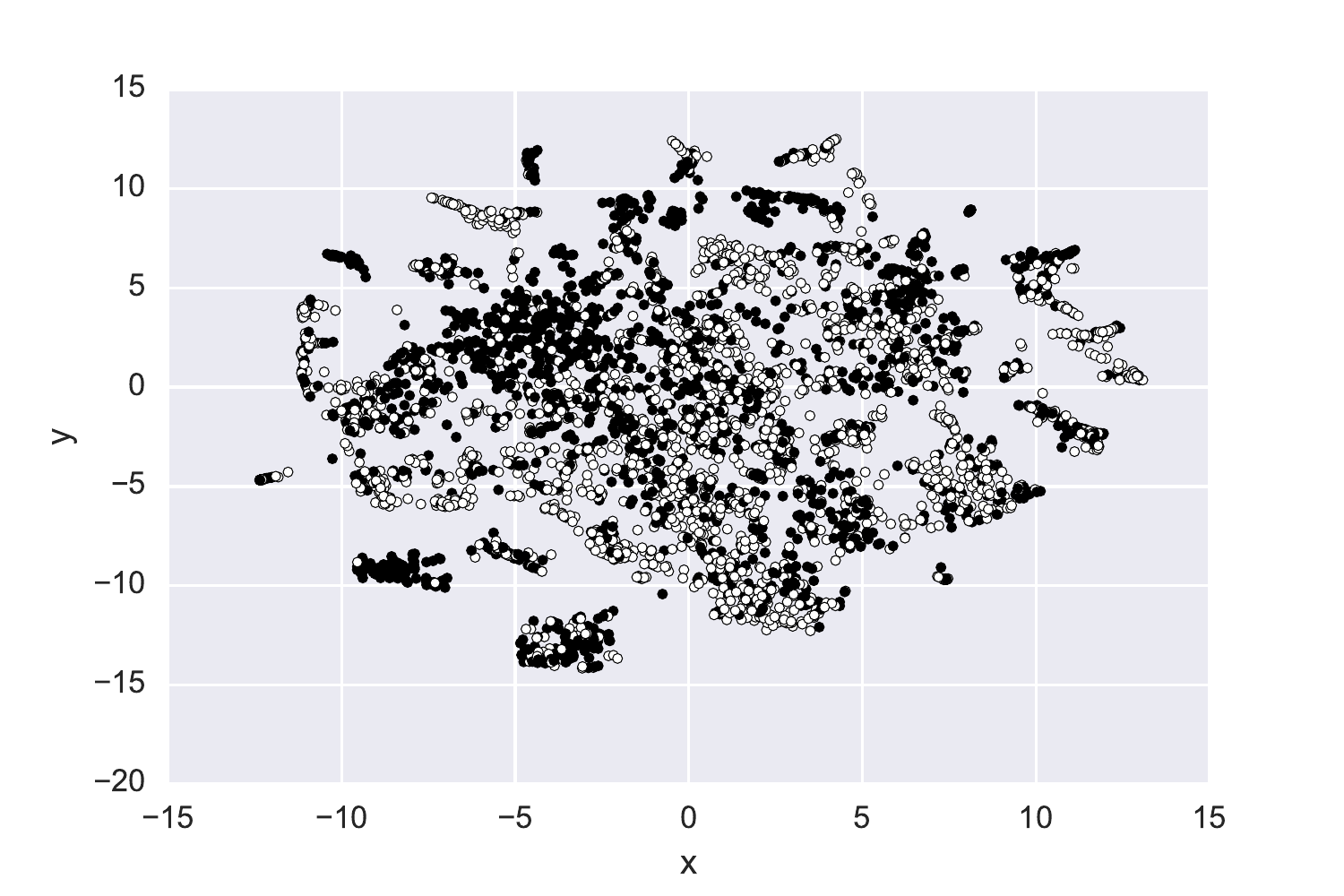}
\caption{A 2-d t-SNE plot of the best performing user embedding (32D). Black and white represent users with above and below median income respectively.}
\label{fig:tsne}
\end{figure}

Our best performing model using graph based features (\emph{Graph}) achieves an accuracy of $50.44\%$ in the occupational classification task. In income prediction, the MAE is $9,048$ and Pearson's correlation is $0.63$. This implies that graph embeddings carry meaningful information about user's socioeconomic attributes making them an effective user representation. The graph embedding features perform consistently better than the textual features (\emph{Topics}) for the majority of the predictive models on both tasks except for the LR model. This confirms our first hypothesis that information from the network structure of a user is indicative of socioeconomic attributes. Figure~\ref{fig:tsne} shows a 2-d t-SNE plot of the best performing user embedding, where we observe many distinct ``communities'' of low and high income users that appear together. This confirms our assumption about the homophilic nature of the Twitter network.

The combination of user embeddings and topics (\emph{Graph+Topics}) outperforms either feature set used individually. More specifically, our \emph{GPC-Graph+Topics} model significantly outperforms (t-test, $p<0.01$) the previous state-of-the-art method, \emph{GPC-Topics} introduced in \citet{jobs2015} on occupational classification. Moreover, our \emph{SVR-Graph+Topics} model significantly outperforms ($p<0.001$) the best baseline method, i.e. \emph{SVM+Topics}. This confirms our second hypothesis that network structure and linguistic information are complementary.

The above findings shed light on the homophilic behaviour of users on Twitter. That might have further implications on user behaviour when selecting friends and forming social networks online. Our results suggest that a stronger bias might exist towards selecting friends with common socioeconomic backgrounds in contrast to common topics of interest and that needs to be explored further.

Non-linear models (i.e. SVM, SVR, GPC, GPR) achieve better results in inferring user socioeconomic attributes than linear (LR) models. While in the occupational classification task our best performing model is GPC, the best model in income inference is the SVR instead of GPR. This implies that model selection is important in these tasks.

%\item Overall predictive performance of our best models indicates the difficulty of predicting income and occupational class attributes. This shows that there is still a lot room for improvement on these two particular tasks. 

An analysis of the errors in the occupational classification task shows that most misclassifications come from adjacent classes. For example, users in classes 1, 3 and 4 are mistakenly classified as class 2. This happens because adjacent classes contain related occupations. However, we notice less dispersion of errors caused by other classes misclassified as class 2 when we use graph embeddings and the combination of graph embeddings and topics. This might be explained by the homophily of users' networks captured by graph embeddings.

\renewcommand\arraystretch{1.0}
\begin{table}[!t]
\centering
\small
\begin{tabular}{|l|c|c|}
\hline
\multicolumn{3}{|c|}{\cellcolor{gray!35}{\bf Income}}\\
\hline
\cellcolor{gray!35} {\bf Method}	& \cellcolor{gray!35} {\bf MAE (\pounds)} & \cellcolor{gray!35} {\bf $\mathbf{\rho}$}\\
\hline
\multicolumn{3}{|c|}{\cellcolor{gray!35}{\bf \citet{income2015}}}\\ 
\hline
LR-Topics 	&	10573 	&	.50	\\
SVR-Topics 	&	9528	& 	.59	\\
GPR-Topics 	&	9883 	&	.60\\	
\hline
\multicolumn{3}{|c|}{\cellcolor{gray!35}{\bf \citet{Hasanuzzaman2017}}}\\ 
\hline
LR-Temporal Or. 	&	10850 	&	.45	\\
GP-Temporal Or. 	&	10235 	&	.51\\	
\hline
\multicolumn{3}{|c|}{\cellcolor{gray!35}{\bf Ours}}\\ 
\hline
LR-Graph	&	10811&	.50 	\\
SVM-Graph	&	{\bf 9048\ddag} & .62	\\
GP-Graph	&	9532 &  .63	\\
\hline
LR-Graph+Topics		&	10326& .54 \\
SVM-Graph+Topics	&	9072\ddag& {\bf.64} \\
GP-Graph+Topics	&	9488 & {\bf.64} \\		
\hline
\end{tabular}
\caption{Mean Absolute Error (MAE) and Pearson's correlation coefficient ($\rho$) between actual and predicted income. $\ddag$ denotes statistical significant different (t-test, $p<0.001$) method to \emph{SVM-Topics}.}
\label{t:results_income}
\end{table}

\section{Conclusions}
\label{sec:conclusions}
We presented a method to reliably predict user occupational class and income on Twitter. Information from a user's social network is represented by graph embeddings \cite{Perozzi2014,Tang2015} and is used to train predictive models. %User graph embeddings are fixed-length, dense vectors  representing users in which similar users have similar vectors in a high-dimensional Euclidean space. We train predictive models for income and occupational class from these embeddings.
% We define predictive tasks where user information, represented by their graph embedding vectors, is used to train supervised models for predicting income and occupational class. 
To the best of our knowledge, this work is the first to introduce graph embeddings for automatically inferring socioeconomic characteristics. We also demonstrated that the information extracted from the user's social network and their language use are complementary. That combination significantly improves predictive performance. Finally, our proposed models achieve state-of-the-art results in two standard datasets of income and occupational class, significantly outperforming previous methods.

%In the future, we plan to enrich the user social graph with content based information such as linguistic features. That could be achieved by weighting the edges of the graph using user content similarity. User content similarity can be computed as the cosine of the angle of the topic vectors of users (or another embedding approach such as doc2vec \cite{Le2014}). Another interesting direction would be to use deep learning methods. For example, a neural network can be used to automatically learn the interplay between language and network feature representations. Furthermore, Long-Short Term Memory \cite{Hochreiter1997} or Convolutional Neural Networks \cite{Kim2014ab} could be employed to automatically learn new features directly from users' tweets. Finally, we would like to apply our proposed method in other social networks, although that requires data become publicly available. 

\bibliographystyle{ACM-Reference-Format}
\bibliography{sigproc} 

\end{document}